\documentclass[lettersize,journal]{IEEEtran}
\usepackage{amsmath,amsfonts}
\usepackage{algorithmic}
\usepackage{algorithm}
\usepackage{array}
\usepackage[caption=false,font=normalsize,labelfont=sf,textfont=sf]{subfig}
\usepackage{textcomp}
\usepackage{bm}
\usepackage{multirow}
\usepackage{stfloats}
\usepackage{url}
\usepackage{verbatim}
\usepackage{graphicx}
\usepackage{cite}
\usepackage[pagebackref=true,breaklinks=true,colorlinks,bookmarks=false,citecolor=blue,linkcolor=blue]{hyperref}

\begin{document}

\title{
DeepEraser: Deep Iterative Context Mining \par for Generic Text Eraser}

\author{
 Hao Feng,
 Wendi Wang,
 Shaokai Liu,
 Jiajun Deng, \par
 Wengang Zhou*,~\IEEEmembership{Senior Member,~IEEE},
 and~Houqiang Li*,~\IEEEmembership{Fellow,~IEEE}
 
\IEEEcompsocitemizethanks{
\IEEEcompsocthanksitem Hao Feng, Wendi Wang, Shaokai Liu, Wengang Zhou, and Houqiang Li are with the CAS Key Laboratory of Technology in Geo-spatial Information Processing and Application System, Department of Electronic Engineering and Information Science, University of Science and Technology of China, Hefei, 230027, China.
Hao Feng is also with Zhangjiang Lab.
E-mail: \{haof, wendiwang, liushaokai\}@mail.ustc.edu.cn; \{zhwg, lihq\}@ustc.edu.cn
\IEEEcompsocthanksitem Jiajun Deng is with The University of Adelaide, Australian Institute for Machine Learning. E-mail: jiajun.deng@adelaide.edu.au
\IEEEcompsocthanksitem *Corresponding authors: Wengang Zhou and Houqiang Li.
}}

\maketitle

\begin{abstract}
In this work, we present DeepEraser, an effective deep network for generic text removal. DeepEraser utilizes a recurrent architecture that erases the text in an image via iterative operations. Our idea comes from the process of erasing pencil script, where the text area designated for removal is subject to continuous monitoring and the text is attenuated progressively, ensuring a thorough and clean erasure. Technically, at each iteration, an innovative erasing module is deployed, which not only explicitly aggregates the previous erasing progress but also mines additional semantic context to erase the target text. Through iterative refinements, the text regions are progressively replaced with more appropriate content and finally converge to a relatively accurate status. 
Furthermore, a custom mask generation strategy is introduced to improve the capability of DeepEraser for adaptive text removal, as opposed to indiscriminately removing all the text in an image. Our DeepEraser is notably compact with only 1.4M parameters and trained in an end-to-end manner. To verify its effectiveness, extensive experiments are conducted on several prevalent benchmarks, including SCUT-Syn, SCUT-EnsText, and Oxford Synthetic text dataset. The quantitative and qualitative results demonstrate the effectiveness of our DeepEraser over the state-of-the-art methods, as well as its strong generalization ability in custom mask text removal. The codes and pre-trained models are available at \href{https://github.com/fh2019ustc/DeepEraser}{{\tt https://github.com/fh2019ustc/DeepEraser}}
\end{abstract}


\section{Introduction}
\IEEEPARstart{T}ext removal in digital images has attracted increasing research attention in the computer vision community.
Its main objective is to remove texts from images and replace them with appropriate content that blends with the surrounding context.
In a world increasingly conscious of data privacy,
this technique has many valuable applications, 
including concealing sensitive information like addresses, license plate numbers, identification numbers, and so on. 
Moreover, it has proved to be useful in multifaceted applications, such as intelligent education~\cite{bojorque2020academic}, text editing~\cite{wu2019editing,yang2020swaptext,krishnan2023textstylebrush,shimoda2021rendering}, image retrieval~\cite{aker2017analyzing,gao2015database,dong2018cross}, and augmented reality translation~\cite{petter2011automatic,rose2016word,syahidi2018bandoar}.

\begin{figure}[t]
	\centering
	\includegraphics[width=1\columnwidth]{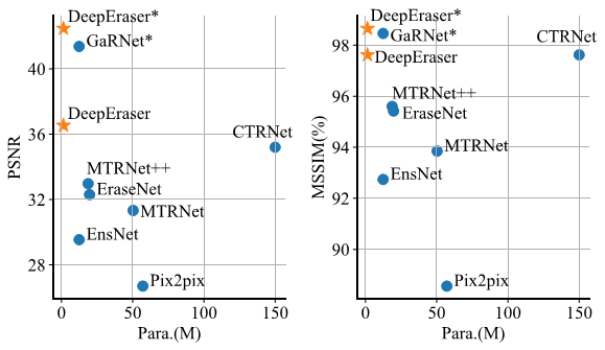}
	\caption{Quantitative metrics for different methods on the SCUT-EnsText benchmark~\cite{zdenek2020erasing}. 
    ``*'' denotes that the predicted text-free images preserve the non-text regions of inputs, while the other methods use the model outputs directly. 
    Our DeepEraser presents the best performance while enjoying the fewest number of model parameters.
	}
	\label{fig:1}
\end{figure}

Thanks to the advancements in deep learning, the field of text removal has witnessed remarkable progress in recent years. 
Among them, many solutions~\cite{zhang2019ensnet,liu2020erasenet,tursun2019mtrnet,tursun2020mtrnet++,susladkar2022tpfnet,liu2022don} are based on GAN~\cite{goodfellow2020generative}. 
These approaches either adopt the generator-discriminator architecture~ and leverage adversarial loss~\cite{zhang2019ensnet,tursun2019mtrnet,liu2020erasenet,susladkar2022tpfnet,tursun2020mtrnet++,liu2022don} to encourage the model to produce more plausible results. 
However, the generated images often suffer from artifacts, incomplete erasure, \emph{etc}.
To address this issue, 
some methods~\cite{tang2021stroke,bian2022scene,lee2022surprisingly,du2023progressive,du2023modeling} utilize text stroke detection and perform background inpainting in identified regions. 
Although they report encouraging performance, the demands for a fine-grained text region mask inevitably introduce additional complexity and uncertainty, 
and text stroke prediction is another challenging problem remained to be solved~\cite{wang2021semi,xu2021rethinking}. 
Recently, some solutions~\cite{wang2023real,lyu2022psstrnet,du2023progressive} based on iterative erasure have been proposed.
Nevertheless, these approaches neglect the exploitation of semantic context in the regions surrounding the target text to be erased, leading to suboptimal results.

In this work, 
we present DeepEraser,
an end-to-end generic deep network for text removal.
We draw the inspiration from the process of erasing pencil
script,
where the area designated for text erasure is subject to continuous monitoring and the text gradually fades away, 
achieving clean and complete erasure.
Technically, DeepEraser takes a recurrent structure that erases the text in an image via iterative context mining and context updates.
Specifically,
at each iteration, a shared erasing module is developed to explicitly aggregate the previous erasing progress and then update the predicted text-free image.
As the iteration times increase,
the network progressively attends to the surrounding context
of the target texts,
mining more semantic information for further erasure.
Consequently, the text regions are progressively inpainted with more appropriate content and blend with their surrounding context.
Note that all the update operations are performed on a fixed high resolution identical to the input text image.
This is different from the intuitive strategy~\cite{zhang2019ensnet,liu2020erasenet,liu2022don,lee2022surprisingly} which refines the result with an image pyramid where early errors could be accumulated and affect the performance.
Finally, we obtain a text-free image with more plausible details.

Additionally, our DeepEraser exhibits several strengths. 
Firstly,
our DeepEraser is lightweight with only 1.4M parameters. 
This is attributed to our compact network design.
Particularly, the erasing module is neat and shares weights across iterations.
As shown in Fig.~\ref{fig:1},
compared with existing approaches~\cite{wolf2006object,zhang2019ensnet,tursun2019mtrnet,liu2020erasenet,tursun2020mtrnet++,liu2022don,lee2022surprisingly},
our DeepEraser achieves the best performance on the SCUT-EnsText benchmark~\cite{zdenek2020erasing} while enjoying the fewest number of parameters.
Secondly, unlike the existing methods~\cite{zhang2019ensnet,liu2020erasenet,tursun2020mtrnet++,tang2021stroke,lyu2022psstrnet,susladkar2022tpfnet,liu2022don,lee2022surprisingly} that rely on a variety of complex loss functions, our training objective is simple: 
we solely compute the $L_1$ distance between the predicted text-free image and the ground truth one. 
Thirdly,
to enhance the ability for adaptive text removal instead of simply removing all text regions in an image~\cite{nakamura2017scene,zhang2019ensnet,liu2020erasenet,susladkar2022tpfnet},
we introduce a custom mask generation strategy.
Concretely,
during training, we randomly select the text instances to generate a mask that indicates the text to be removed;
during inference, 
we can customize the mask with existing text detectors~\cite{baek2019character,liao2020real,zhang2021adaptive} or by simply scribbling on any desired text regions to be erased.

To evaluate the effectiveness of our proposed DeepEraser,
we conduct comprehensive experiments on several prevalent benchmark datasets, 
including SCUT-EnsText~\cite{liu2020erasenet}, SCUT-Syn~\cite{zhang2019ensnet}, and Oxford Synthetic text dataset~\cite{gupta2016synthetic}.
The quantitative and qualitative results demonstrate the superiority of our DeepEraser over the state-of-the-art methods and its strong generalization ability to custom mask text removal. 

In summary, we make three-fold contributions:
\begin{itemize}
\item We propose DeepEraser, an end-to-end deep network for text removal. It takes a recurrent structure that erases the text via iterative context mining and context updates.

\item We introduce a custom mask generation strategy to facilitate adaptive text removal, and present an elegant design on the network and training objective.

\item We conduct extensive experiments to validate the merits of our method, 
and demonstrate significant improvements on several prevalent benchmark datasets.

\end{itemize}

\section{Related Work}
Over the years, many approaches have been proposed to tackle the text removal task.
Traditional methods~\cite{telea2004image,khodadadi2012text,wagh2015text} rely on handcrafted features with complex algorithms for image restoration. 
These methods are suitable for simple scenarios, 
but exhibit limited efficacy if the images have perspective distortion and complicated backgrounds. 
Recently, deep learning-based methods have shown impressive results and require less manual intervention. 
In the following, we categorize the learning-based methods into two groups.

\subsection{One-stage Methods}
The one-stage approaches are commonly based on an end-to-end model.
Pix2pix~\cite{wolf2006object} leverages conditional adversarial network~\cite{mirza2014conditional} to learn the mapping from input images to output ones, 
which can be applied to scene text removal. 
STE~\cite{nakamura2017scene} presents the first DNN-based model for scene text removal. 
It utilizes a single-scaled sliding-window-based neural network that takes small patches of the image as input, 
allowing for the removal of text on a small scale.
However, this approach may sacrifice the overall consistency of the restored output.
Currently, GAN~\cite{goodfellow2020generative} has been adopted by methods~\cite{zhang2019ensnet,liu2020erasenet,tursun2019mtrnet,tursun2020mtrnet++,susladkar2022tpfnet,liu2022don} for text removal.
EnsNet~\cite{zhang2019ensnet} employs a GAN-based network and adopts four carefully designed loss functions to further enhance performance. 
EraseNet~\cite{liu2020erasenet} and MTRNet++~\cite{tursun2020mtrnet++} introduce a coarse-refinement architecture and an additional branch to help locate the text. 
CTRNet~\cite{liu2022don} introduces a text perception head for text region positioning. It explores both low-frequency structure and high-level context features to guide the process of text erasure and background restoration. 
It utilizes Transformer~\cite{vaswani2017attention} to capture local features and establish the long-term relationship among pixels globally.
Additionally, 
GaRNet~\cite{lee2022surprisingly} uses gated attention to focus on the text stroke and the surrounding regions.
A region-of-interest generation methodology is introduced, 
which focuses on only the text region to train the model more efficiently.

\begin{figure*}[t]
	\centering
	\includegraphics[width=2\columnwidth]{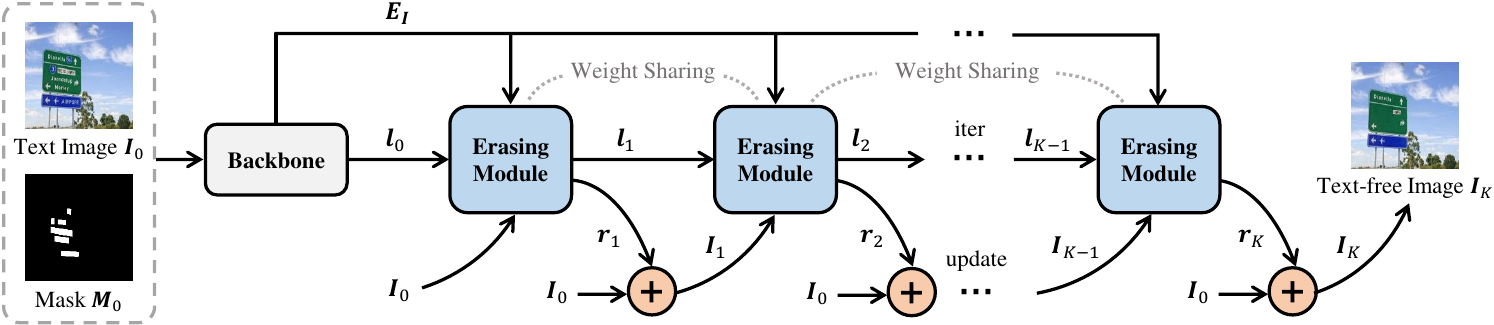}
	\caption{An overview of DeepEraser for text removal. 
    Given a text image $\bm{I}_0$ and a binary mask $\bm{M}_0$ indicating the regions for text removal, we first extract the feature through a CNN-based backbone. Then, a shared erasing module refines the estimated text-free image across $K$ iterations.
    At the $k^{th}$ iteration, it explicitly aggregates the previous erasing progress and outputs the residual image $\bm{r}_k$ to update the erasing result $\bm{I}_k$.
    After $K$ iterations, we obtain the final predicted text-free image $\bm{I}_K$.
	}
	\label{fig:framework}
\end{figure*}

\subsection{Multi-stage Methods}
Several methods decompose the text removal task into two sub-problems, \emph{i.e.,} text detection and background inpainting. 
Text detection can be manual or automatic.
Among them,
Qin~\emph{et al.}~\cite{qin2018automatic} introduce a cGAN-based architecture with one encoder and two decoders for the segmentation of the foreground text stroke and background inpainting. 
MTRNet~\cite{tursun2019mtrnet} proposes a mask-based two-stage method,
which shows the significance of masks for the improvement of performance. 
Tang~\emph{et al.}~\cite{tang2021stroke} explores cropping the text regions and predicting the text strokes, 
then the network inpaints the stroke pixels with the appropriate content. 
Zdenek~\emph{et al.}~\cite{zdenek2020erasing} propose a weak supervision method consisting of a pre-trained scene text detector~\cite{wang2019shape} and a pre-trained GAN-based inpainting module~\cite{zheng2019pluralistic} for effective text removal.

Recently, some solutions~\cite{wang2023real,lyu2022psstrnet,du2023progressive} based on iterative erasure have been proposed. Typically, 
PSSTRNet introduced a novel mask update mechanism for progressively refining text masks and implemented an adaptive fusion approach to optimize the use of outcomes across various iterations. PERT~\cite{du2023progressive} addresses background integrity and text erasure exhaustivity by using explicit erasure guidance for stroke-level modifications and a balanced multi-stage erasure process. However, these methods fail to effectively utilize the semantic context of the areas surrounding the text targeted for erasure.

\smallskip
Although the field of text removal has witnessed rapid progress in recent years, 
there remain unsolved problems such as artifacts, incomplete erasure, and insufficient semantic extraction. 
In this work, we introduce an iterative erasing strategy,
aiming to progressively mine the image context and then replace the target text with more appropriate content.

\section{Approach}
An overview of the proposed DeepEraser is presented in Fig.~\ref{fig:framework}.
Given a text image $\bm{I}_0 \in \mathbb{R}^{H \times W \times 3}$ to be processed and a binary mask map $\bm{M}_0 \in \mathbb{R}^{H \times W \times 1}$ indicating the regions for text removal, we aim to estimate an image $\bm{I}_K \in \mathbb{R}^{H \times W \times 3}$ where the target text in $\bm{I}_0$ are removed while the pixel values in other regions are maintained.
Here, $H$ and $W$ are the image height and width.
Our DeepEraser erases the target text of image $\bm{I}_0$ with $K$ iterations.
At each iteration, a shared erasing module is developed to explicitly aggregate the previous erasing
progress and then update the output text-free image.
Through iterative erasure operations, the designated text regions are progressively replaced with appropriate content, converging to a thorough and clean erasure.

The workflow of our method can be distilled down to three stages, including 
(i) custom mask generation to produce a binary map $\bm{M}_0$, indicating the text regions to be erased; (ii) feature extraction from the image $\bm{I}_0$ and the mask $\bm{M}_0$; and (iii) iterative text erasing that removes the target text progressively. 
We separately detail them below.

\subsection{Custom Mask Generation}~\label{sec:mask}
To facilitate adaptive text removal,
during the training stage, we randomly select the text instances in $\bm{I}_0$ rather than removing all text instances in the image.
Specifically, each text instance undergoes filtering with a probability $\alpha$.
Then, our training objective is to remove only the selected text.
An example of a mask $\bm{M}_0$ is present in Fig.~\ref{fig:framework}.
It is noteworthy that such a strategy can also enhance the robustness of the network for attending to the indicated text regions.

During inference, we can customize the mask $\bm{M}_0$ based on the off-the-shelf text detectors~\cite{baek2019character,liao2020real} or by scribbling on any desired text regions.
The image $\bm{I}_0$ and the obtained mask $\bm{M}_0$ are then fed into DeepEraser for text removal.

\subsection{Feature Extraction}
\begin{figure}[t]
	\centering
	\includegraphics[width=1\columnwidth]{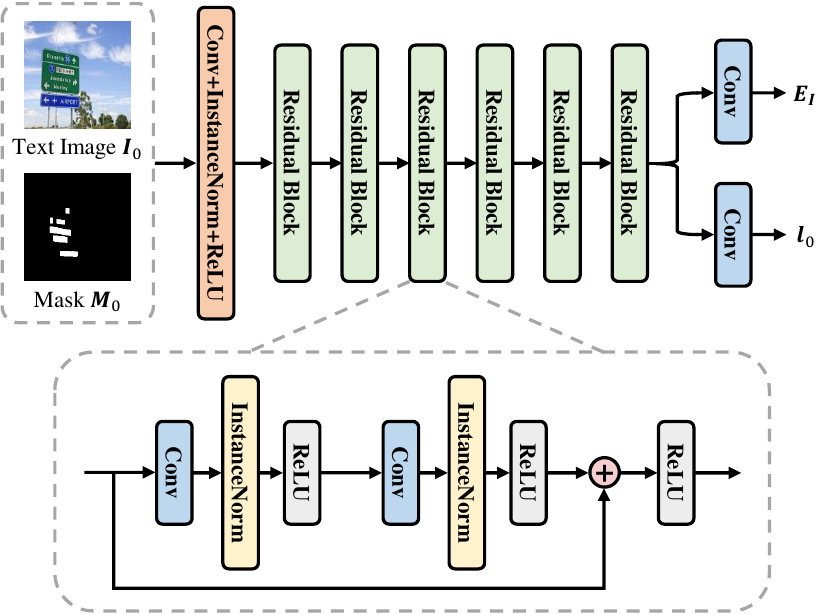}
        \caption{Architecture of backbone for feature extraction.}
	\label{fig:backbone}
\end{figure}

Given a text image $\bm{I}_0 \in \mathbb{R}^{H \times W \times 3}$ and a mask $\bm{M}_0 \in \mathbb{R}^{H \times W \times 1}$, we first concatenate them along the channel dimension, and feed the output into a conventional CNN backbone for feature extraction.
As shown in Fig.~\ref{fig:backbone}, 
we present the specific architecture of the backbone network for feature extraction. 
Our backbone consists of six residual blocks~\cite{he2016deep}.
In order to generate a finer feature map for the subsequent text erasure, 
we do not involve any downsampling operations here.
Next, we cascade two parallel convolutional layers to produce the context feature $\bm{E}_I \in \mathbb{R}^{H \times W \times D}$ and the initial latent feature $\bm{l}_0 \in \mathbb{R}^{H \times W \times D}$, where the channel dimension $D$ is set to 64 by default. 
It is noteworthy that our backbone network is lightweight and has only 0.9M parameters.

\subsection{Iterative Text Erasing}
The core component in our DeepEraser is the erasing module, 
which iteratively refines the current removal results.
As shown in Fig.~\ref{fig:framework},
we estimate a sequence of residual images $\{\bm{r}_1,\bm{r}_2,...,\bm{r}_K\}$, 
where $K$ is the number of iterations and $\bm{r}_k \in \mathbb{R}^{H \times W \times 3}$ denotes the residual image used to update the previous removal results as follows,
\begin{equation}\label{update_image}
    \bm{I}_k=\bm{I}_0+\bm{r}_k, k=\{1,2,...,K\}.
\end{equation}

\noindent
At the $k^{th}$ iteration, 
the erasing module takes (i) context feature $\bm{E}_I$, (ii) previous estimated text-free image $\bm{I}_{k-1}$, and (iii) latent feature $\bm{l}_{k-1}$ as input, 
and outputs the updated latent feature $\bm{l}_{k}$ and current residual image $\bm{r}_k$. Note that the weights of erasing module are shared across iterations.

\begin{figure}[t]
	\centering
	\includegraphics[width=0.95\columnwidth]{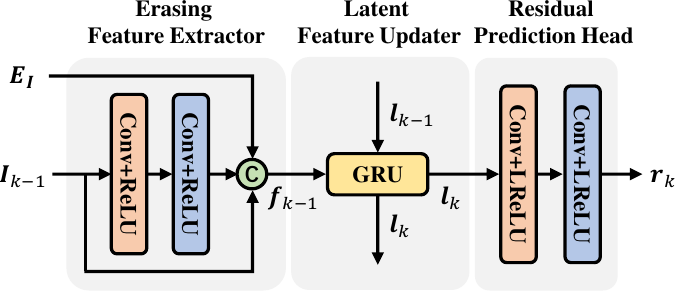}
	\caption{An illustration of the $k^{th}$ iteration in the erasing module.
         It takes (1) context feature $\bm{E}_I$, (2) previously estimated text-free image $\bm{I}_{k-1}$, and (3) latent feature $\bm{l}_{k-1}$ as input, and outputs the updated latent $\bm{l}_{k}$ and current residual image $\bm{r}_k$.
	}
	\label{fig:update}
\end{figure}

As shown in Fig.~\ref{fig:update}, 
we divide the erasing module into three blocks: (i) feature extractor, (ii) latent feature updater, and (iii) residual prediction head. We detail them below.

\smallskip
\textbf{Erasing Feature Extractor.}
The erasing feature extractor \textit{explicitly} encodes the previous erasing progress.
Concretely,
given the text-free image $\bm{I}_{k-1}$ predicted at the $(k-1)^{th}$ iteration, 
we first apply two convolutional layers on it, and then concatenate the output with $\bm{I}_{k-1}$ itself.
Thereafter,
the output feature map is further concatenated with the context feature $\bm{E}_I$ to produce the feature map $\bm{f}_{k-1}$.

\smallskip
\textbf{Latent Feature Updater.}
Different from the obtained $\bm{f}_{k-1}$, 
the latent feature $\bm{l}_{k-1}$ \textit{implicitly} models the previous erasing progress.
The core component of this block is a gated activation unit based on the GRU cell~\cite{cho2014properties}, 
with the fully connected layers replaced with convolutional layers~\cite{teed2020raft}.
At the $k^{th}$ iteration,
it processes the input $\bm{f}_{k-1} \in \mathbb{R}^{H \times W \times D}$ as well as latent feature $\bm{l}_{k-1} \in \mathbb{R}^{H \times W \times D}$, and outputs the
updated latent feature $\bm{l}_{k} \in \mathbb{R}^{H \times W \times D}$ as follows,
\begin{equation}
   \begin{aligned}
       \textbf{x}_t&=\sigma(Conv_{3 \times 3}([\bm{l}_{k-1},\bm{f}_{k-1}],\bm{W}_x)), \\
       \textbf{y}_t&=\sigma(Conv_{3 \times 3}([\bm{l}_{k-1},\bm{f}_{k-1}],\bm{W}_y)), \\
       \tilde{\bm{l}_k}&=tanh(Conv_{3 \times 3}([\textbf{y}_t\odot \bm{l}_{k-1},\bm{f}_{k-1}], \bm{W}_r)), \\
       \bm{l}_k&=(1-\textbf{x}_t) \odot \bm{r}_{k-1}+\textbf{x}_t \odot \tilde{\bm{l}_k},
       \label{original_form}
   \end{aligned}
\end{equation}
where $\sigma$ stands for the standard sigmoid function, $\odot$ denotes the scalar product operation of matrices, and weight matrices are represented as $\bm{W}$.

\begin{figure*}[t]
	\centering
	\includegraphics[width=1.98\columnwidth]{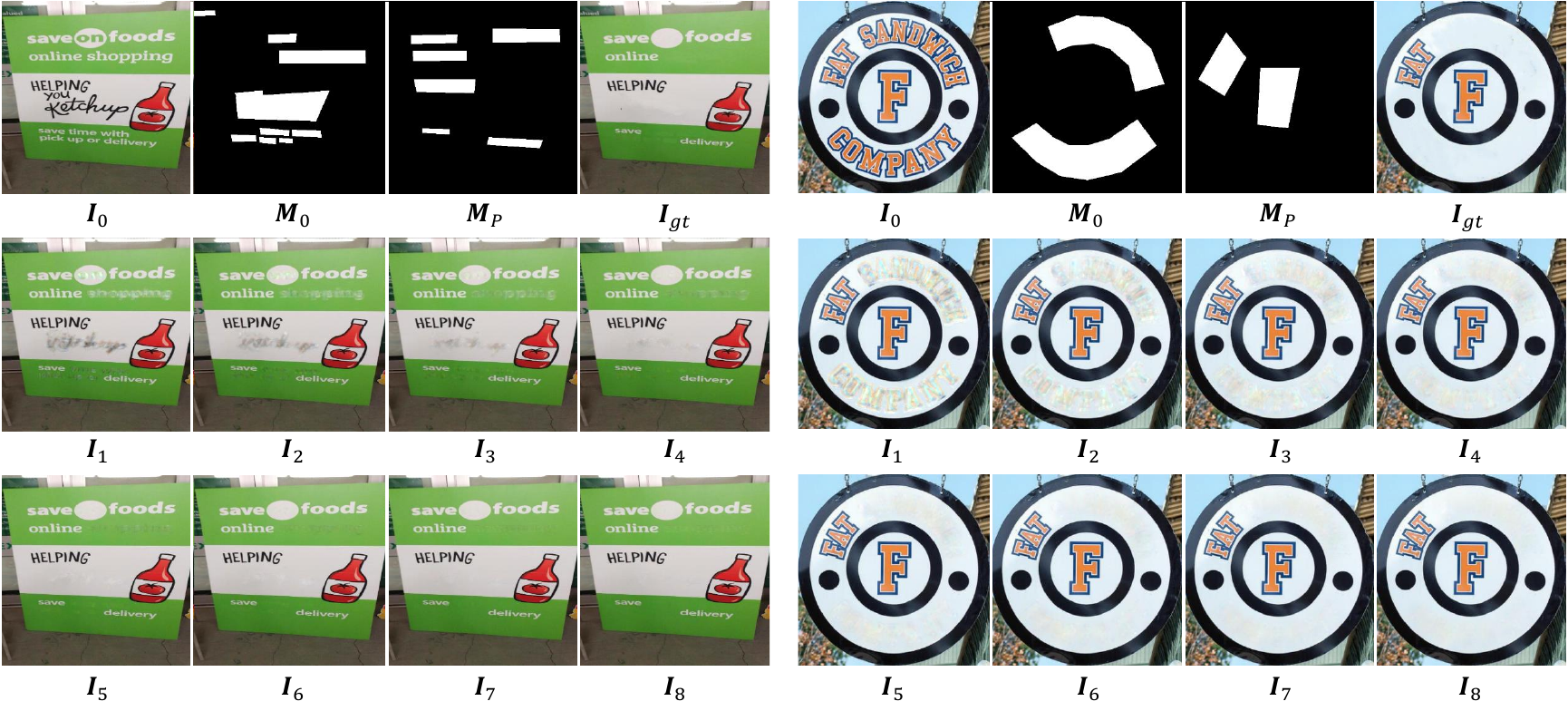}
	\caption{Qualitative results of each iteration in the inference stage on SCUT-EnsText~\cite{zdenek2020erasing}.
    The first row of the two examples presents the input text image $\bm{I}_0$, the input text-erased mask $\bm{M}_0$, the text-preserved mask $\bm{M}_p$, and the ground truth $\bm{I}_{gt}$, respectively. 
    The second and third rows are the predicted text-free images $\bm{I}_k, k=\{1,2,...,8\}$. 
    With the iteration times increasing, the text is erased progressively. 
	}
	\label{fig:iteration}
\end{figure*}

\begin{figure*}[t]
	\centering
	\includegraphics[width=1.98\columnwidth]{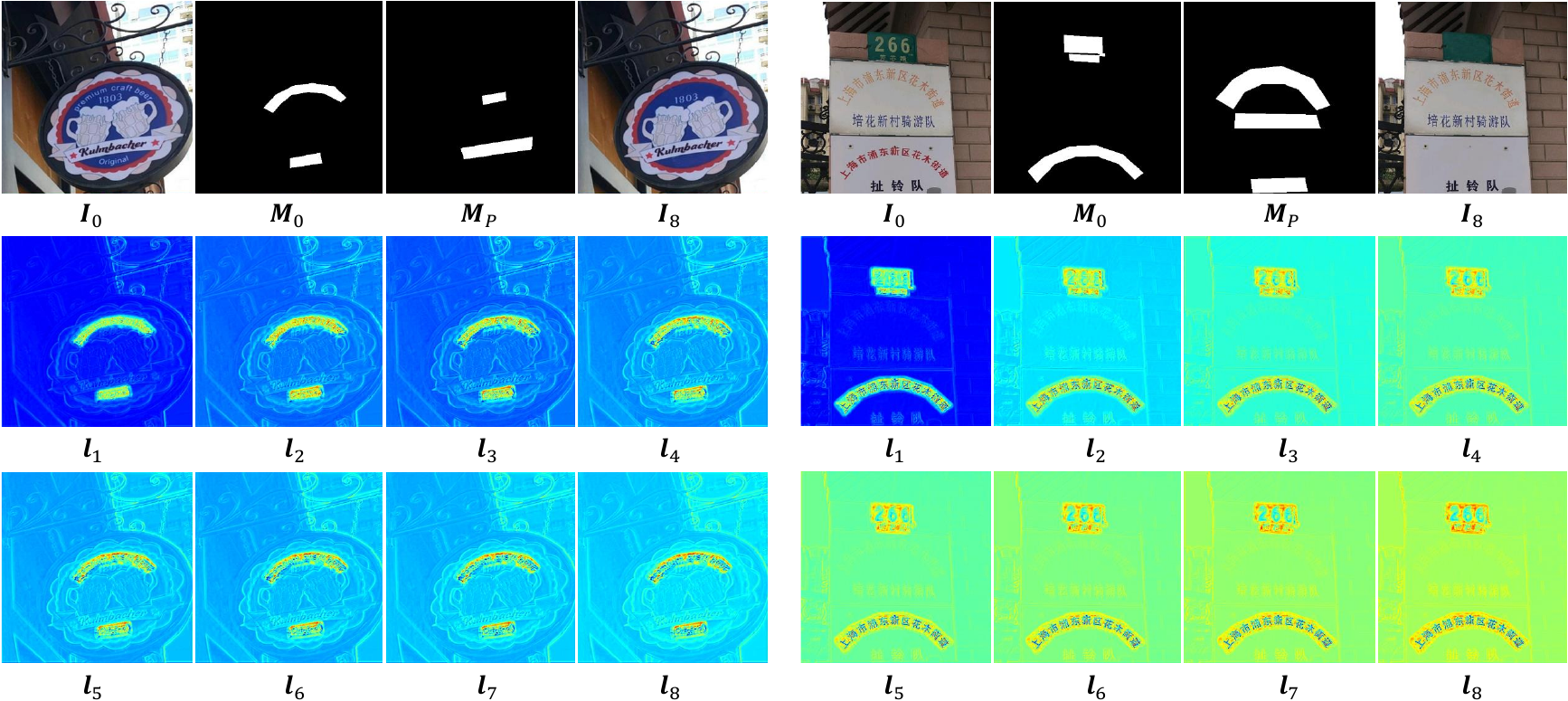}
	\caption{Visualization of the latent feature $\bm{l}_k$ at each iteration in the inference stage on SCUT-EnsText~\cite{zdenek2020erasing}.
    The first row of two examples shows the input text image $\bm{I}_0$, the input text-erased mask $\bm{M}_0$, the text-preserved mask $\bm{M}_p$, and the output text-free image $\bm{I}_8$, respectively.
    The second and third rows present the visualization of latent feature $\bm{l}_k, k=\{1,2,...,8\}$.
    As the iteration rounds increase,
    the network increasingly attends to the surrounding context of the indicated texts,
    mining additional semantic information to inpaint the text regions.
	}
	\label{fig:fea}
\end{figure*}

In Fig.~\ref{fig:fea}, we observe that as the iteration times increase, the network
increasingly attends to the surrounding context of the indicated texts, indicating that more semantic information is utilized to erase the target text.

\setlength{\tabcolsep}{2.1mm}   
\begin{table*}[t]
 	\caption{Ablation experiments on the SCUT-EnsText dataset~\cite{zdenek2020erasing}, including
    the settings of the erasing module, weight sharing, supervision iterations, and input mask generation. 
    The default settings in our final model are drawn with \underline{underscores}.
    }
        \renewcommand\arraystretch{1.1}
	\centering
        \begin{tabular}{c|cc|cccccc|ccc|c}
        \hline
        \multirow{2}{*}{Ablation} & \multicolumn{2}{c|}{\multirow{2}*{Setting}} & \multicolumn{6}{c}{Image-Eval} & \multicolumn{3}{|c|}{Detection-Eval} & \multirow{2}{*}{Para. (M)} \\
         & & & PSNR $\uparrow$ & MSSIM $\uparrow$ & MSE $\downarrow$ & AGE $\downarrow$ & pEPs $\downarrow$ & pCEPS $\downarrow$ & $\mathcal{P}$ $\downarrow$ & $\mathcal{R}$ $\downarrow$ & $\mathcal{F}$ $\downarrow$\\
        \hline
        \multirow{5}{*}{Erasing module} & (a) & None & 21.52 & 90.02 & 0.81 & 18.64 & 0.3621 & 0.2231 & 55.4 & 7.7 & 13.5 & \textbf{0.98} \\
         & (b) & $\bm{r}_k \rightarrow \bm{I}_k$ & 35.54 & 97.44 & 0.09 & 1.85 & 0.0119 & 0.0083 & 21.4 & 1.4 & 3.7 & 1.44 \\
         & (c) & w/o $\bm{E}_{\bm{I}}$ & 36.29 & 97.58 & \textbf{0.07} & 1.68 & 0.0096 & 0.0064 & 8.8 & 0.5 & 0.9 & 1.31 \\
         & (d) & w/o $\bm{I}_k$ & 24.59 & 94.04 & 0.44 & 12.52 & 0.1217 & 0.0650 & 39.6 & 3.4 & 6.3 & 1.32 \\
         & (e) & \underline{full module} & \textbf{36.53} & \textbf{97.62} & \textbf{0.07} & \textbf{1.63} & \textbf{0.0091} & \textbf{0.0059} & \textbf{3.6} & \textbf{0.2} & \textbf{0.3} & 1.44 \\
        \hline
        \multirow{2}{*}{Weights sharing} & (f) & w/o & 36.42 & 97.57 & \textbf{0.07} & 1.66 & 0.0095 & 0.0062 & 4.8 & 0.2 & 0.4 & 5.19 \\
         & (e) & \underline{w} & \textbf{36.53} & \textbf{97.62} & \textbf{0.07} & \textbf{1.63} & \textbf{0.0091} & \textbf{0.0059} & \textbf{3.6} & \textbf{0.2} & \textbf{0.3} & \textbf{1.44} \\
        \hline
        \multirow{2}{*}{Supervised iters} & (g) & $\{K\}$ & 36.37 & 97.54 & \textbf{0.07} & 1.67 & 0.0097 & 0.0064 & 4.3 & 0.2 & 0.4 & - \\
         & (e) & \underline{$\{1,2,..,K\}$} & \textbf{36.53} & \textbf{97.62} & \textbf{0.07} & \textbf{1.63} & \textbf{0.0091} & \textbf{0.0059} & \textbf{3.6} & \textbf{0.2} & \textbf{0.3} & \textbf{1.44} \\
         \hline
        \multirow{3}{*}{Mask} & (h) & no mask & 31.83 & 96.17 & 0.23 & 2.56 & 0.0198 & 0.0133 & 55.2 & 11.2 & 18.6 & - \\
         & (i) & all mask & \textbf{36.55} & \textbf{97.66} & \textbf{0.07} & 1.64 & \textbf{0.0091} & \textbf{0.0059} & 5.8 & 0.3 & 0.5 & - \\
         & (e) & \underline{part mask} & 36.53 & 97.62 & \textbf{0.07} & \textbf{1.63} & \textbf{0.0091} & \textbf{0.0059} & \textbf{3.6} & \textbf{0.2} & \textbf{0.3} & \textbf{1.44} \\
        \hline
        \end{tabular}
	\label{tab:ablation}
\end{table*}

\setlength{\tabcolsep}{3.7mm}   
\begin{table*}[t]
 	\caption{Ablation experiments of the iteration number $K$ during \textbf{training} on the SCUT-EnsText dataset~\cite{zdenek2020erasing}. 
    Increasing the iteration number $K$ steadily improves the performance. 
    To strike a balance between accuracy and efficiency, 
    we choose $K=8$ by default. 
    }
        \small
        \renewcommand\arraystretch{1.1}
	\centering
        \begin{tabular}{c|cccccc|ccc|c}
        \hline
        \multirow{2}{*}{Iters} & \multicolumn{6}{c}{Image-Eval} & \multicolumn{3}{|c|}{Detection-Eval} & \multirow{2}{*}{FPS} \\
        & PSNR $\uparrow$ & MSSIM $\uparrow$ & MSE $\downarrow$ & AGE $\downarrow$ & pEPs $\downarrow$ & pCEPS $\downarrow$ & $\mathcal{P}$ $\downarrow$ & $\mathcal{R}$ $\downarrow$ & $\mathcal{F}$ $\downarrow$ & \\
        \hline
        4 & 36.09 & 97.43 & 0.08 & 1.72 & 0.0106 & 0.0071 & 14.0 & 0.9 & 1.6 & \textbf{3.95} \\
        6 & 36.35 & 97.56 & \textbf{0.07} & 1.67 & 0.0097 & 0.0064 & 5.2 & 0.2 & 0.5 & 2.97 \\
        8 & 36.53 & 97.62 & \textbf{0.07} & 1.63 & 0.0091 & 0.0059 & 3.6 & 0.2 & 0.3 & 2.37 \\
        10 & 36.64 & 97.68 & \textbf{0.07} & \textbf{1.60} & 0.0085 & \textbf{0.0054} & 3.7 & 0.2 & 0.3 & 1.95 \\
        12 & \textbf{36.72} & \textbf{97.71} & \textbf{0.07} & \textbf{1.60} & \textbf{0.0084} & \textbf{0.0054} & \textbf{2.8} & \textbf{0.1} & \textbf{0.2} & 1.67 \\
        \hline
        \end{tabular}
	\label{tab:iteration_1}
\end{table*}

\setlength{\tabcolsep}{2.3mm}   
\begin{table*}[t]
 	\caption{Quantitative results at the selected iteration during \textbf{inference} on the SCUT-EnsText dataset~\cite{zdenek2020erasing}. 
    The performance progressively improves and finally converges to a stable state without divergence. 
    To strike a balance between accuracy and efficiency, 
    we set $K=8$. 
    }
        \small
	\centering
        \begin{tabular}{c|ccccccccccccc}
        \hline
        Metrics$\backslash
$iters & 1 & 2 & 3 & 4 & 5 & 6 & 7 & 8 & 9 & 10 & 12 & 16 & 32 \\
        \hline
        PSNR $\uparrow$ & 34.28 & 35.34 & 35.85 & 36.13 & 36.30 & 36.40 & 36.47 & 36.53 & 36.56 & 36.59 & 36.61 & \textbf{36.62} & 36.60 \\
        MSSIM $\uparrow$ & 95.89 & 96.83 & 97.25 & 97.43 & 97.53 & 97.57 & 97.60 & 97.62 & 97.63 & 97.63 & \textbf{97.64} & \textbf{97.64} & 97.63 \\
        FPS $\uparrow$ & \textbf{8.73} & 6.31 & 4.96 & 4.07 & 3.45 & 2.99 & 2.64 & 2.37 & 2.14 & 1.96 & 1.67 & 1.29 & 0.68 \\
        \hline
        \end{tabular}
	\label{tab:iteration_2}
\end{table*}

\setlength{\tabcolsep}{5mm}
\begin{table}[t]
 	\caption{Ablation experiment about the format of the input mask in training. The performances are evaluated on a new test set with partial text instances to be removed.
    }
        \small
	\centering
        \begin{tabular}{c|ccc}
        \hline
        Mask & PSNR $\uparrow$ & MSSIM $\uparrow$ & AGE $\downarrow$ \\
        \hline
        all mask & 37.36 & 97.88 & 1.46 \\
        part mask & \textbf{37.52} & \textbf{98.01} & \textbf{1.40} \\
        \hline
        \end{tabular}
	\label{tab:generic}
\end{table}

\smallskip
\textbf{Residual Prediction Head.}
The prediction head is followed by the current latent feature $\bm{l}_{k} \in \mathbb{R}^{H \times W \times D}$ to produce the residual image $\bm{r}_k \in \mathbb{R}^{H \times W \times 3}$.
It applies feature projection by two convolutional layers with LeakyReLU activation~\cite{xu2015empirical}.
Then $\bm{r}_k$ is used to update the text-free image (Eq.~\eqref{update_image}).
After $K$ iterations, the final output $\bm{I}_K$ is obtained. 

\subsection{Training Objective}
In contrast to existing text removal methods that employ a variety of complex losses,
our loss function is simple,
defined as the sum of the $L_1$ distance between the ground truth image $\bm{I}_{gt}$ and the predicted text-free one at each iteration:
\begin{equation}
    \mathcal{L} = \sum_{k=1}^K \lambda^{K-k} \left \| \bm{I}_{gt} - \bm{I}_k \right \|_1,
    \label{supervision}
\end{equation}
where $\lambda^{K-k}$ is the weight of the $k^{th}$ iteration which increases exponentially ($\lambda < 1$).

\section{Experiment}
\subsection{Datasets}
We train and evaluate our method on three commonly used benchmark datasets, including SCUT-EnsText~\cite{liu2020erasenet}, SCUT-Syn~\cite{zhang2019ensnet}, and the Oxford Synthetic text dataset~\cite{gupta2016synthetic}.

\smallskip
\textbf{SCUT-EnsText}~\cite{liu2020erasenet} benchmark is an extensive real-world dataset featuring a collection of Chinese and English text images. These images are compiled from a variety of public scene text reading benchmark datasets. It encompasses a total of 3,562 natural images, annotated with over 21,000 text instances. The dataset is divided into a training set, which includes 2,749 images containing 16,460 words, and a test set, consisting of 813 images with 4,864 words.

\smallskip
\textbf{SCUT-Syn}~\cite{zhang2019ensnet} benchmark uses text synthesis technology~\cite{gupta2016synthetic} to generate samples on scene images. It consists of 8,000 training images and 800 test images, all standardized to a resolution of 512$\times$512 pixels.

\smallskip
\textbf{The Oxford Synthetic text dataset}~\cite{gupta2016synthetic} comprises approximately 800,000 synthetic images.
As suggested in GaRNet~\cite{lee2022surprisingly}, the dataset is partitioned into 735,364 images for training and 30,000 for testing purposes, respectively.

\subsection{Metrics}
We evaluate our method according to the conventions in the literature, utilizing both Image-Eval and Detection-Eval metrics for comprehensive evaluation.

\smallskip
\textbf{Image-Eval.}
Image-Eval aims to assess the quality of the predicted text-free image. 
Following previous methods~\cite{zhang2019ensnet,tursun2020mtrnet++,liu2020erasenet,lee2022surprisingly,liu2022don}, 
our evaluation metrics consist of Peak Signal-to-Noise Ratio (PSNR), Multi-scale Structure Similarity Index Measure (MSSIM)~\cite{wang2003multiscale}, Mean Squared Error (MSE), along with
AGE, pEPs, and pCEPS~\cite{zhang2019ensnet}.

\setlength{\tabcolsep}{2.3mm}
\begin{table*}[t]
        \small
 	\caption{Quantitative comparison on \textbf{SCUT-EnsText dataset}~\cite{zdenek2020erasing}. 
    ``\dag'' denotes that the input mask for evaluation is generated with the detected results~\cite{baek2019character}. 
    ``*'' denotes that the predicted images preserve the non-text regions of inputs, while the other methods use the model outputs.
    }
	\centering
        \begin{tabular}{l|cccccc|ccc|c|c}
        \hline
        \multirow{2}{*}{Method} & \multicolumn{6}{c}{Image-Eval} & \multicolumn{3}{|c|}{Detection-Eval} & \multirow{2}{*}{FPS} & \multirow{2}{*}{Para. (M)} \\
         & PSNR $\uparrow$ & MSSIM $\uparrow$ & MSE $\downarrow$ & AGE $\downarrow$ & pEPs $\downarrow$ & pCEPS $\downarrow$ & $\mathcal{P}$ $\downarrow$ & $\mathcal{R}$ $\downarrow$ & $\mathcal{F}$ $\downarrow$ & & \\
        \hline
        Pix2pix~\cite{wolf2006object} & 26.70 & 88.56 & 0.37 & 6.09 & 0.0480 & 0.0227 & 69.7 & 35.4 & 47.0 & 22.5 & 57.1 \\
        STE~\cite{nakamura2017scene} & 25.46 & 90.14 & 0.47 & 6.01 & 0.0533 & 0.0296 & 40.9 & 5.9 & 10.2 & - & - \\
        EnsNet~\cite{zhang2019ensnet} & 29.54 & 92.74 & 0.24 & 4.16 & 0.0307 & 0.0136 & 68.7 & 32.8 & 44.4 & \textbf{34.25} & 12.4 \\
        MTRNet~\cite{tursun2019mtrnet} & 31.33 & 93.85 & 0.13 & 3.53 & 0.0256 & 0.0086 & 71.2 & 42.1 & 52.9 & 13.87 & 50.3 \\
        MTRNet++~\cite{tursun2020mtrnet++} & 32.97 & 95.60 & 0.20 & 2.49 & 0.0186 & 0.0118 & 58.9 & 15.0 & 24.0 & 6.33 & 18.7 \\
        EraseNet~\cite{liu2020erasenet} & 32.30 & 95.42 & 0.15 & 3.02 & 0.0160 & 0.0090 & 53.2 & 4.6 & 8.5 & 12.48 & 19.7 \\
        Ours & \textbf{36.53} & \textbf{97.62} & \textbf{0.07} & \textbf{1.63} & \textbf{0.0091} & \textbf{0.0059} & \textbf{3.6} & \textbf{0.2} & \textbf{0.3} & 2.37 & \textbf{1.4} \\
        
        \hline
        CTRNet\dag~\cite{liu2022don} & 35.20 & 97.36 & 0.09 & - & - & - & 38.4 & 1.4 & 2.7 & 1.06 & 150.0 \\
        Ours\dag & \textbf{35.84} & \textbf{97.48} & \textbf{0.08} & \textbf{1.71} & \textbf{0.0101} & \textbf{0.0064} & \textbf{31.6} & \textbf{0.5} & \textbf{1.0} & \textbf{2.37} & \textbf{1.4} \\
        
        \hline
        CTRNet*~\cite{liu2022don} & 37.20 & 97.66 & 0.07 & - & - & - & - & - & - & 1.06 & 150.0 \\
        GaRNet*~\cite{lee2022surprisingly} & 41.37 & 98.46 & - & 0.64 & - & - & 15.5 & 1.0 & 1.8 & \textbf{33.70} & 12.4 \\
        Ours* & \textbf{42.47} & \textbf{98.65} & - & \textbf{0.59} & - & - & \textbf{9.9} & \textbf{0.2} & \textbf{0.3} & 2.37 & \textbf{1.4} \\
        \hline
        \end{tabular}
	\label{tab:EnsText}
\end{table*}

\smallskip
\textbf{Detection-Eval.}
Detection-Eval considers how much text has been erased while ignoring the image quality.
As recommended in previous works~\cite{liu2020erasenet,tursun2020mtrnet++,lee2022surprisingly,liu2022don}, 
CRAFT~\cite{baek2019character} is served as the auxiliary text detector.
Then, precision ($\mathcal{P}$), recall ($\mathcal{R}$), and F-score ($\mathcal{F}$)~\cite{ma2018arbitrary} are calculated, expected to be zero after perfect text removal.

\setlength{\tabcolsep}{0.9mm}
\begin{table}[t]
 	\caption{Quantitative comparison on the \textbf{SCUT-Syn dataset}~\cite{liu2020erasenet}. 
    }
        \footnotesize
        \renewcommand\arraystretch{1.1}
	\centering
        \begin{tabular}{l|cccccc}
        \hline
        Method & PSNR $\uparrow$ & MSSIM $\uparrow$ & MSE $\downarrow$ & AGE $\downarrow$ & pEPs $\downarrow$ & pCEPS $\downarrow$\\
        \hline
        Pix2pix~\cite{wolf2006object} & 26.76 & 91.08 & 0.27 & 5.4678 & 0.0473 & 0.0244 \\
        STE~\cite{nakamura2017scene} & 25.40 & 90.12 & 0.65 & 9.4853 & 0.0553 & 0.0347 \\
        EnsNet~\cite{zhang2019ensnet} & 37.36 & 96.44 & 0.21 & 1.7300 & 0.0069 & 0.0020 \\
        MTRNet~\cite{tursun2019mtrnet} & 29.71 & 94.43 & - & - & - & - \\
        MTRNet++~\cite{tursun2020mtrnet++} & 34.55 & 98.45 & 0.04 & - & - & - \\
        EraseNet~\cite{liu2020erasenet} & 38.32 & 97.67 & 0.02 & 1.5982 & 0.0048 & 0.0004 \\
        CTRNet~\cite{liu2022don} & 41.28 & 98.50 & 0.02 & - & - & - \\
        Ours & \textbf{42.10} & \textbf{98.65} & \textbf{0.01} & \textbf{1.0816} & \textbf{0.0019} & \textbf{0.0002} \\
        \hline
        \end{tabular}
	\label{tab:Syn}
\end{table}

\subsection{Implementation Details}

We implement our DeepEraser in PyTorch~\cite{paszke2017automatic}.
All modules are initialized from scratch with random weights and then optimized in an end-to-end manner.
During training, Adam optimizer~\cite{kingma2014adam} with an initial learning rate $10^{-4}$ is adopted,
and we use the 1cycle policy~\cite{smith2019super} for learning rate decay.
The images are cropped to $256\times 256$ for training and keep their original resolution (\emph{i.e.}, $512\times 512$) for testing.
We train our model for 200 epochs on the SCUT-Syn~\cite{zhang2019ensnet} and SCUT-EnsText~\cite{liu2020erasenet} dataset with a batch size of 2.
For the Oxford Synthetic text dataset~\cite{gupta2016synthetic}, 
due to its large amount of data, we train for only 20 epochs. 
Two NVIDIA GeForce GTX 3090Ti GPUs are leveraged in our experiments.
We set the default iteration number $K$ to 8, 
and the iteration number during inference is the same as that in the training stage.
The probability $\alpha$ for mask generation during training is set as $40\%$.
We set the hyperparameter $\lambda$ to 0.85 in Eq.~\eqref{supervision}.

\subsection{Ablation Studies}
In this section,
we perform extensive ablation studies to validate the core components of our DeepEraser.
All ablated versions are trained on the SCUT-EnsText dataset~\cite{liu2020erasenet} following~\cite{liu2022don}.
Several intriguing properties are observed.

\smallskip
\textbf{Erasing Module.}
We first validate our core erasing module by removing it and cascading a CNN-based prediction head behind the backbone.
Experiments (a) and (e) in Tab.~\ref{tab:ablation} demonstrate that
without the erasing module, PSNR and MSSIM drop by 41.09\% and 7.79\%, respectively.

By default, at the $k^{th}$ iteration, the prediction head produces the residual image $\bm{r}_k$ to update the text-free image. 
Then, we ablate a version by directly estimating the text-free image $\bm{I}_k$. 
In Tab.~\ref{tab:ablation}, experiments (b) and (e) show that predicting residual images works better.
This could be ascribed that, the prediction of residual image relieves the network from predicting the known non-text regions, 
allowing it to concentrate on the recovery of the text regions.

As shown in Fig.~\ref{fig:update}, at the $k^{th}$ iteration, 
the erasing module takes 
(1) context feature $\bm{E}_I$, (2) previous estimated text-free image $\bm{I}_{k-1}$, and (3) latent feature $\bm{l}_{k-1}$ as input.
We ablate $\bm{E}_I$ and $\bm{I}_{k-1}$ separately. 
In Tab.~\ref{tab:ablation},
experiments (c), (d), and (e) demonstrate that $\bm{E}_I$ has a weak effect on performance, 
while $\bm{I}_{k-1}$ can greatly improve the erasing quality.
The reason is that the explicit feature extraction from $\bm{I}_{k-1}$ encodes the previous erasing progress which is utilized to guide the further erasure at the current iteration. 

\setlength{\tabcolsep}{4mm}
\begin{table}[t]
        \small
 	\caption{Quantitative comparison on the \textbf{Oxford Synthetic text dataset}~\cite{gupta2016synthetic}.
    In the evaluation, 
    all the predicted images preserve non-text regions of inputs. 
    }
	\centering
        \begin{tabular}{l|ccc}
        \hline
        Method & PSNR $\uparrow$ & MSSIM $\uparrow$ & AGE $\downarrow$  \\
        \hline
        EnsNet~\cite{zhang2019ensnet} & 39.74 & 97.94 & 0.77 \\
        MTRNet~\cite{tursun2019mtrnet} & 40.03 & 97.69 & 0.80 \\
        MTRNet++~\cite{tursun2020mtrnet++} & 40.64 & 97.94 & 0.73 \\
        EraseNet~\cite{liu2020erasenet} & 42.98 & 98.75 & 0.56 \\
        GaRNet~\cite{lee2022surprisingly} & 43.64 & 98.64 & 0.55 \\
        Ours & \textbf{44.85} & \textbf{98.97} & \textbf{0.50} \\
        \hline
        \end{tabular}
	\label{tab:Oxford}
\end{table}

\smallskip
\textbf{Iteration Mechanism.}
We ablate the iteration number $K$ in training. 
Tab.~\ref{tab:iteration_1} proves the importance of sufficient iteration times. 
Like the process of erasing the pencil script,
the more times the erasures, the cleaner the paper will be.
However, as we can see, larger iteration times also bring lower efficiency. 
To balance the computational cost, 
we choose $K=8$ in our final model, 
which provides a great speedup benefit while also enjoying a good performance. 

To provide a specific view of the erasing process, 
we visualize the results of each iteration in the inference stage.
As shown in Fig.~\ref{fig:iteration}, with the iteration rounds increase, 
the text decays progressively.
The main recovery lies in the top 1$\sim$4 iterations, while the
later iterations fine-tune the performance.
Furthermore, in Tab.~\ref{tab:iteration_2}, 
we investigate the performance at the selected iteration during inference. 
As we can see, the performance improves steadily with more iterations and finally converges. 
Note that the performance does not diverge even when the iteration number $K$ is increased to 32.
These quantitative
and qualitative results demonstrate the effectiveness and 
robustness of our DeepEraser.

\begin{figure}[t]
	\centering
	\includegraphics[width=0.99\columnwidth]{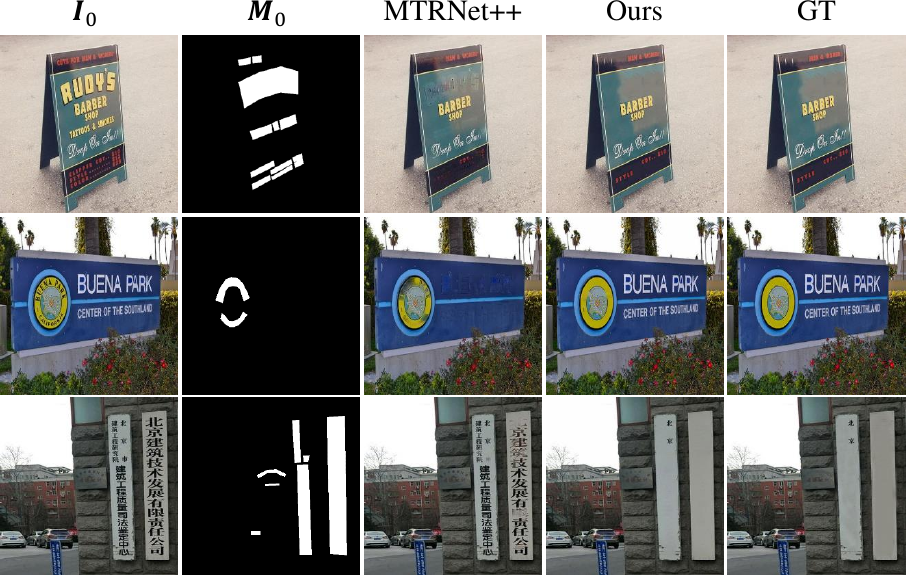}
	\caption{Qualitative comparison of DeepEraser and MTRNet++~\cite{tursun2020mtrnet++} on custom mask text removal on SCUT-EnsText~\cite{liu2020erasenet}. 
    Each row shows the input text image $\bm{I}_0$, the input text-erased mask $\bm{M}_0$, the results of MTRNet++~\cite{tursun2020mtrnet++} and our methods, and the ground truth. 
	}
    \vspace{-0.1in}
	\label{figure:part_mask}
\end{figure}

\begin{figure*}[t]
	\centering
	\includegraphics[width=2.04\columnwidth]{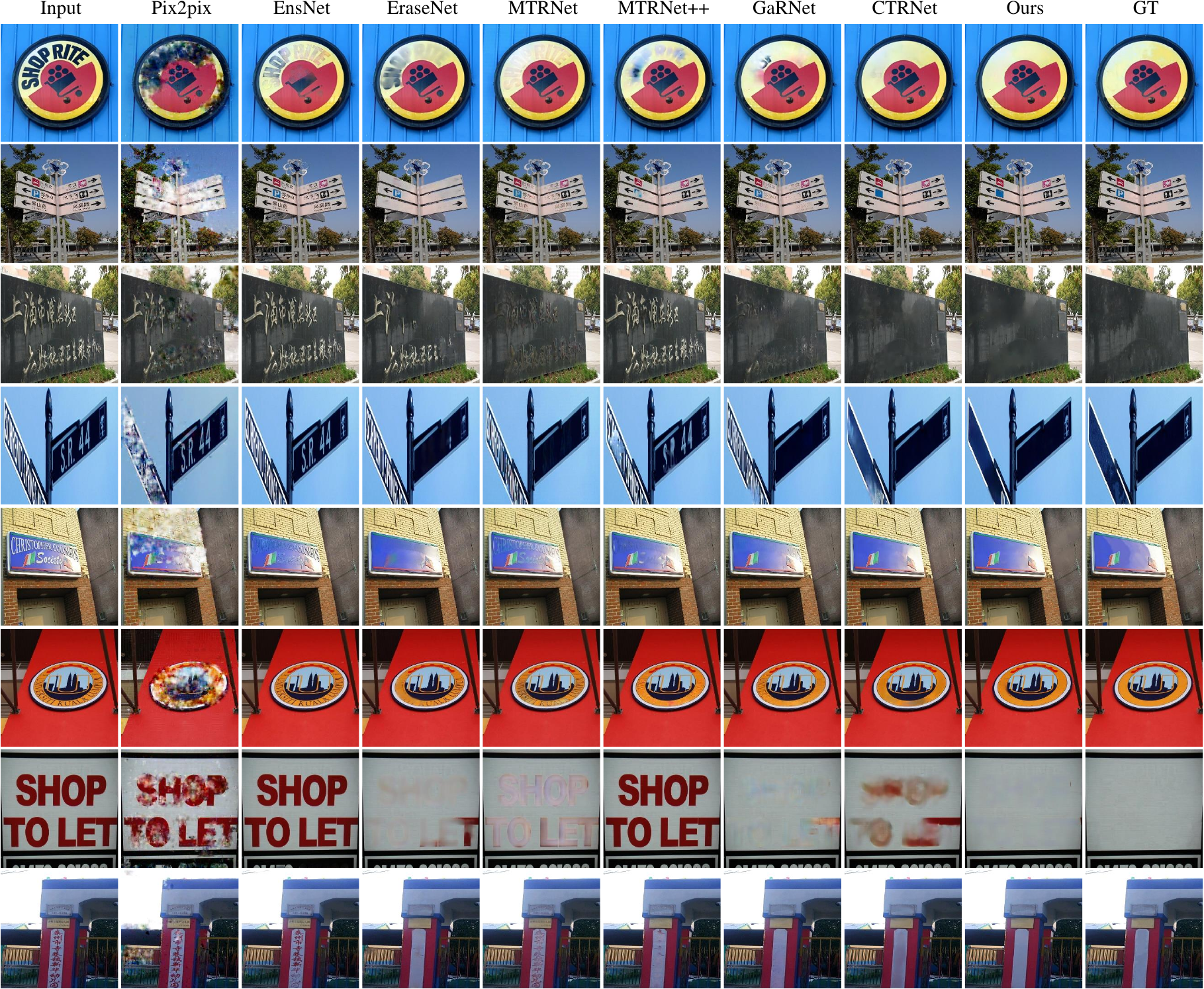}
	\caption{Qualitative comparison on the SCUT-EnsText~\cite{zdenek2020erasing}.
    For each comparison, we show the input text image, the predicted text-free results of the compared methods: Pix2pix~\cite{wolf2006object},  EnsNet~\cite{zhang2019ensnet}, EraseNet~\cite{liu2020erasenet}, 
    MTRNet~\cite{tursun2019mtrnet}, MTRNet++~\cite{tursun2020mtrnet++},  GaRNet~\cite{lee2022surprisingly}, CTRNet~\cite{liu2022don}, our DeepEraser, and the ground truth, from left to right.
    Zoom in for a better view.
	}
	\label{fig:result2}
\end{figure*}

We visualize the latent feature at each iteration in the inference stage,
to illustrate how our method works.
Fig.~\ref{fig:fea} shows that when the iteration times increase, the network pays increasing attention to the surrounding context of the indicated texts,
achieving the mining of additional semantic information for replacing the target text.
In this way, the text regions are progressively erased with the more appropriate content that matches the surrounding environment.

\smallskip
\textbf{Weight Sharing.}
An important design of our DeepEraser is to share the weights of the erasing module across the $K$ iterations. 
Experiments (f) and (e) in Tab.~\ref{tab:ablation} study this design. 
When each module learns a separate set of weights individually, performances become slightly worse, 
% the obtained PSNR and SSIM drop by 0.3\% and 0.05\% respectively, 
but the number of parameters increases significantly. 
This can be attributed to the increased training difficulty of a large
model size, which is also verified in other tasks~\cite{peng2020deep}.

\smallskip
\textbf{Supervised Iterations.}
By default, we calculate the sum of the $L_1$ distance between the predicted text-free image and its ground truth at each iteration. Subsequently, we investigate a variant wherein the loss is computed exclusively at the final iteration. The effectiveness of our supervision setup is verified by experiments (g) and (e) present in Table~\ref{tab:ablation}.
This can potentially be attributed to the application of weighted supervision at each iteration, which facilitates a progressively clearer optimization objective as the iterations advance, thereby enhancing the learning of erasure operations.

\smallskip
\textbf{Mask Generation.}
To facilitate adaptive text removal,
we randomly select the text instances to generate the input mask during training,
while keeping all the text instances during testing (see Sec.~\ref{sec:mask}). 
Then, we study a variant where all text instances are retained for mask generation during training (see experiment (i) in Tab.~\ref{tab:ablation}).
Compared with our default part mask training strategy (see experiment (e) in Tab.~\ref{tab:ablation}),
they show comparable performances.

Then, we construct a new test set with partial text instances to be removed based on the SCUT-EnsText~\cite{liu2020erasenet} dataset. 
Each text instance is selected with a probability $\alpha$
to generate the input mask $\bm{M}_0$ and ground truth $\bm{I}_{gt}$. 
Interestingly, as shown in Tab.~\ref{tab:generic},
performances are better when training with the part mask.
This improvement can be primarily attributed to the fact that our strategy enhances the robustness of the network, by
attending to the indicated text regions rather than all text in an image.
The results are consistent with our motivation to enhance the ability for adaptive text removal.

Additionally, 
we conduct the experiment that merely takes the text image as input (see the experiment (h) in Tab.~\ref{tab:ablation}).
The results indicate that incorporating mask $\bm{M}_0$ as input yields a considerable performance improvement.

\begin{figure*}[t]
	\centering
	\includegraphics[width=2.04\columnwidth]{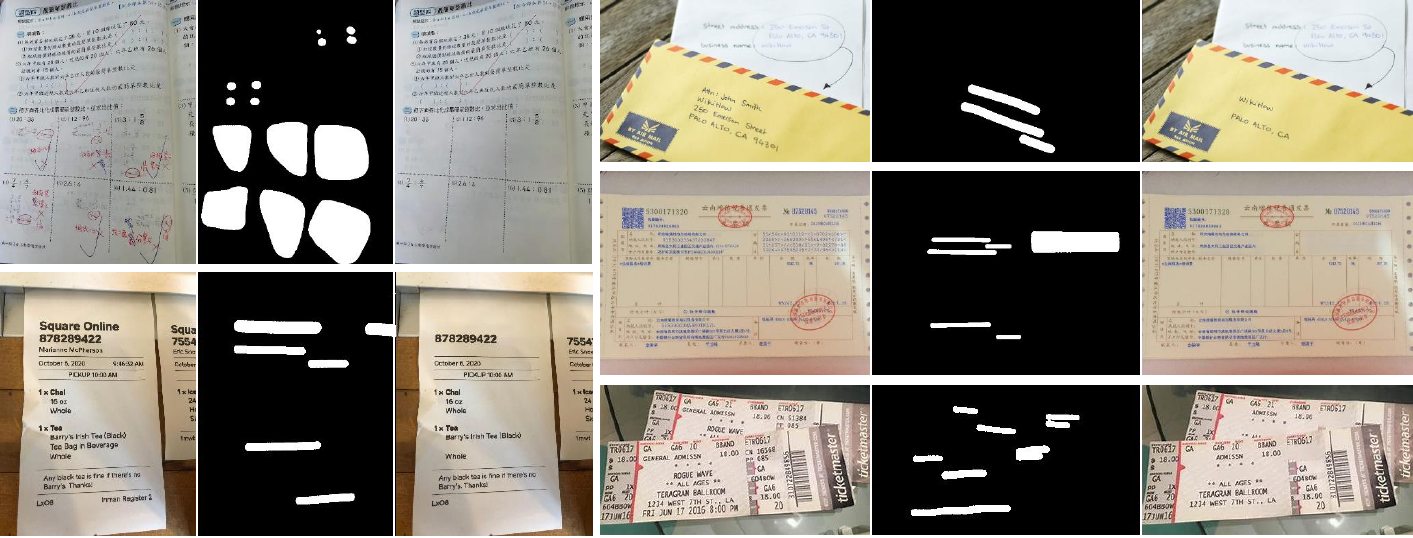}
	\caption{Qualitative results of our DeepEraser for custom mask text removal. Each image triplet shows the input text image, the hand-drawn text-erased mask, and the predicted text-free image. DeepEraser effectively removes the target text and retains the rest.
	}
	\label{figure:generic}
\end{figure*}

To intuitively illustrate the ability of DeepEraser in custom mask text erasure, 
we visualize the results in several application scenarios, 
including intelligent education and privacy protection.
Here all input masks are hand drawn.
As shown in Fig.~\ref{figure:generic},
the target text in the input images is well removed.

\subsection{Comparison with State-of-the-art Methods}
\smallskip
\textbf{Quantitative Comparison.}
In this section, 
we conduct experiments to compare our method with the state-of-the-art methods on three benchmark datasets.
We employ the same evaluation code as CTRNet~\cite{liu2022don} and GaRNet~\cite{lee2022surprisingly}. 
The quantitative results on SCUT-EnsText~\cite{liu2020erasenet}, SCUT-Syn~\cite{zhang2019ensnet}, and Oxford Synthetic text dataset~\cite{gupta2016synthetic} are shown in Tab.~\ref{tab:EnsText}, Tab.~\ref{tab:Syn}, and Tab.~\ref{tab:Oxford}, respectively. 
Note that the results of GaRNet~\cite{lee2022surprisingly} only preserve non-text regions of the input. 
For a fair comparison, 
we re-evaluated our approach in this way.
The results demonstrate that our DeepEraser outperforms the existing advanced methods.
Moreover, our method requires the fewest number of parameters (1.4M).
Note that since our DeepEraser operates iterations on a fixed high resolution,
its efficiency is not exceptionally high. The FPS performance is comparable to CTRNet~\cite{liu2022don} (see Tab.~\ref{tab:EnsText}).

\smallskip
\textbf{Qualitative Comparison.}
We first compare the qualitative results of our DeepEraser with other methods on the SCUT-EnsText dataset~\cite{liu2020erasenet}. 
As shown in Fig.~\ref{fig:result2},
both Pix2pix~\cite{wolf2006object} and EnsNet~\cite{zhang2019ensnet} struggle to handle complex text images. 
Some results generated by EraseNet~\cite{liu2020erasenet}, MTRNet~\cite{tursun2019mtrnet} and MTRNet++~\cite{tursun2020mtrnet++} contain artifacts and discontinuities. 
GaRNet~\cite{lee2022surprisingly} and CTRNet~\cite{liu2022don} also suffer from artifacts and inaccurate erasure regions. In contrast,
our DeepEraser replaces the text regions with more appropriate content and obtains better qualitative results across diverse scenarios.

In addition, since MTRNet++~\cite{tursun2020mtrnet++} studies the 
custom mask text removal,
we further provide a comparison in Fig.~\ref{figure:part_mask}. 
As we can see,
the results of MTRNet++~\cite{tursun2020mtrnet++} exhibit some over-removal and incomplete removal, while the predicted text-free images of our DeepEraser are more plausible.

\section{Conclusion}
In this work, 
we present DeepEraser, an effective and lightweight deep network for generic text removal.
DeepEraser utilizes a novel recurrent architecture that progressively erases the target text in an image through successive iterations.
Within each iteration, a compact erasing module is employed, 
which mines the context around the designated areas and then inpaints them with more appropriate content.
Through iterative refinement,
the text is progressively erased, finally producing a text-free image with more plausible local details.
Extensive experiments are conducted on several prevalent benchmark datasets. 
The quantitative and qualitative results verify the merits of DeepEraser over advanced methods as well as its strong generalization ability in custom mask text removal.

{
	\bibliographystyle{IEEEtran}
	\bibliography{IEEEabrv,eibib}
}

\vfill

\end{document}